\documentclass[sigconf,screen,nonacm]{acmart}
\setcopyright{none}
\renewcommand\footnotetextcopyrightpermission[1]{}
\AtBeginDocument{%
  }

\usepackage{booktabs} % for professional tables
\usepackage{microtype}
\usepackage{graphicx}

\usepackage{subcaption}

\usepackage{tcolorbox}
\usepackage{amsmath}

\usepackage{amssymb}
\usepackage{mathtools}
\usepackage{amsthm}
\usepackage{multirow}
\usepackage{array}
\usepackage{cleveref}
\usepackage{pifont}

\hypersetup{
    pdftitle={Visualizing the Invisible: Generative Visual Grounding Empowers Universal EEG Understanding in MLLMs},
    pdfauthor={Jun-Yu Pan, Yansen Wang, Enze Zhang, Bao-Liang Lu, Wei-Long Zheng, Dongsheng Li}
}

\begin{document}

\title{Visualizing the Invisible: Generative Visual Grounding Empowers Universal EEG Understanding in MLLMs}

% \author{Jun-Yu Pan\textsuperscript{1,\textdagger}, Yansen Wang\textsuperscript{2,*}, Enze Zhang\textsuperscript{1}, Bao-Liang Lu\textsuperscript{1}, Wei-Long Zheng\textsuperscript{1,*}, Dongsheng Li\textsuperscript{2}}
% \affiliation{%
%     \institution{\textsuperscript{1}Shanghai Jiao Tong University \quad \textsuperscript{2}Microsoft Research Asia}
% }
% \email{{panjunyu,zez-626,bllu,weilong}@sjtu.edu.cn} \email{{yansenwang,dongsli}@microsoft.com}
\author{Jun-Yu Pan}
\authornotemark[2]
\affiliation{
  \institution{Shanghai Jiao Tong University}
  \city{Shanghai}
  \country{China}
}
\email{panjunyu@sjtu.edu.cn}

\author{Yansen Wang}
\authornotemark[1]
\affiliation{
  \institution{Microsoft Research Asia}
    \city{Shanghai}
  \country{China}
}
\email{yansenwang@microsoft.com}

\author{Enze Zhang}
\affiliation{
  \institution{Shanghai Jiao Tong University}
  \city{Shanghai}
  \country{China}
}
\email{zez-626@sjtu.edu.cn}

\author{Bao-Liang Lu}
\affiliation{
  \institution{Shanghai Jiao Tong University}
  \city{Shanghai}
  \country{China}
}
\email{bllu@sjtu.edu.cn}

\author{Wei-Long Zheng}
\authornotemark[1]
\affiliation{
  \institution{Shanghai Jiao Tong University}
  \city{Shanghai}
  \country{China}
}
\email{weilong@sjtu.edu.cn}

\author{Dongsheng Li}
\affiliation{
  \institution{Microsoft Research Asia}
    \city{Shanghai}
  \country{China}
}
\email{dongsli@microsoft.com}
\settopmatter{printacmref=false}
\renewcommand\footnotetextcopyrightpermission[1]{}

\makeatletter
\def\@authorsaddresses{}
\makeatother
\renewcommand{\shortauthors}{Pan et al.}

\begin{abstract}
    Leveraging the universal representations of pre-trained Large Language Models (LLMs) and Multimodal Large Language Models (MLLMs) has emerged as a promising paradigm for enhancing the capability of brain foundation models. Constrained by the severe scarcity of visually-evoked EEG datasets, existing foundation models predominantly focus on LLMs and compromise by aligning neural signals solely with abstract text---a lossy translation that inevitably discards the fine-grained, perceptual details encoded in brain activity. In this work, we propose \textbf{G}enerative \textbf{V}isual \textbf{G}rounding (\textbf{GVG}), a framework that visualizes the invisible. Rather than forcing neural signals into text, we employ an EEG-to-Image generative model as a "visual translator" to hallucinate instance-specific proxy images for non-visual EEG. This strategy provides structured visual contexts that allow MLLMs to apply their visual priors to interpret clinical states. To validate this core hypothesis, we first establish a robust image-only alignment across two distinct MLLM backbones (GVG-X-Omni and GVG-Janus). This purely visual paradigm is already competitive: even our lightweight discrete instantiation, GVG-X-Omni, matches 1.7B-parameter text-aligned baselines while tuning merely 170M parameters on top of a frozen 7B backbone (~10$\times$ fewer trainable parameters). Building on this result, we further extend GVG-Janus via a trimodal (Image+Text) alignment. Extensive experiments suggest a consistent multimodal complementarity: textual alignment provides categorical semantic anchors, while visual proxies enrich the neural representations with fine-grained perceptual details. This combination consistently improves over our single-modality variants, enhancing both EEG understanding and visual generative capabilities. At a comparable parameter scale, GVG-Janus remains competitive with NeuroLM and improves over it on several benchmarks. These results suggest that visual proxy grounding can serve as an effective complement to textual alignment for universal EEG understanding.
\end{abstract}

% CCS metadata is omitted in the arXiv/preprint version.
\maketitle
\begingroup
\renewcommand{\thefootnote}{\fnsymbol{footnote}}
\footnotetext[1]{Corresponding authors.}
\footnotetext[2]{Work done during Jun-Yu's internship at Microsoft Research Asia.}
\endgroup

\section{Introduction}
The emergence of Large Language Models (LLMs) and Multimodal Large Language Models (MLLMs) has revolutionized AI~\cite{achiam2023gpt,team2023gemini,wang2024qwen2}, inspiring the development of foundation models to learn universal representations for non-invasive brain signals such as electroencephalography (EEG)~\cite{wang2023brainbert,yi2023learning,jiang2024large,wang2024cbramod,fang2025neuript}. While the scaling law of EEG foundations~\cite{banville2025scaling} has revealed a positive correlation between EEG data volume and downstream performance, unlike text or images, high-quality EEG data are expensive to collect and constrained by stringent privacy concerns, resulting in limited data availability that severely restricts the development of higher-performing EEG foundation models. Consequently, further scaling training purely on such limited EEG data inevitably leads to severe overfitting.

Recent approaches have attempted to bridge EEG and other modalities, starting from mapping neural signals to the text space~\cite{jiang2024neurolm,kim2024eeg,lu2025unimind,liu2025echo}. On the one hand, this alignment unlocks the capability of multi-task learning, making it possible to utilize all kinds of labels to train a unified decoder. On the other hand, bringing both modalities into the shared representation space harnesses the vast semantic knowledge encapsulated in pre-trained LLMs, thus circumventing the data bottleneck~\cite{huh2024platonic}. 

\begin{figure*}[t]
    \centering
    \begin{subfigure}{0.49\textwidth}
        \centering
        \includegraphics[width=\linewidth]{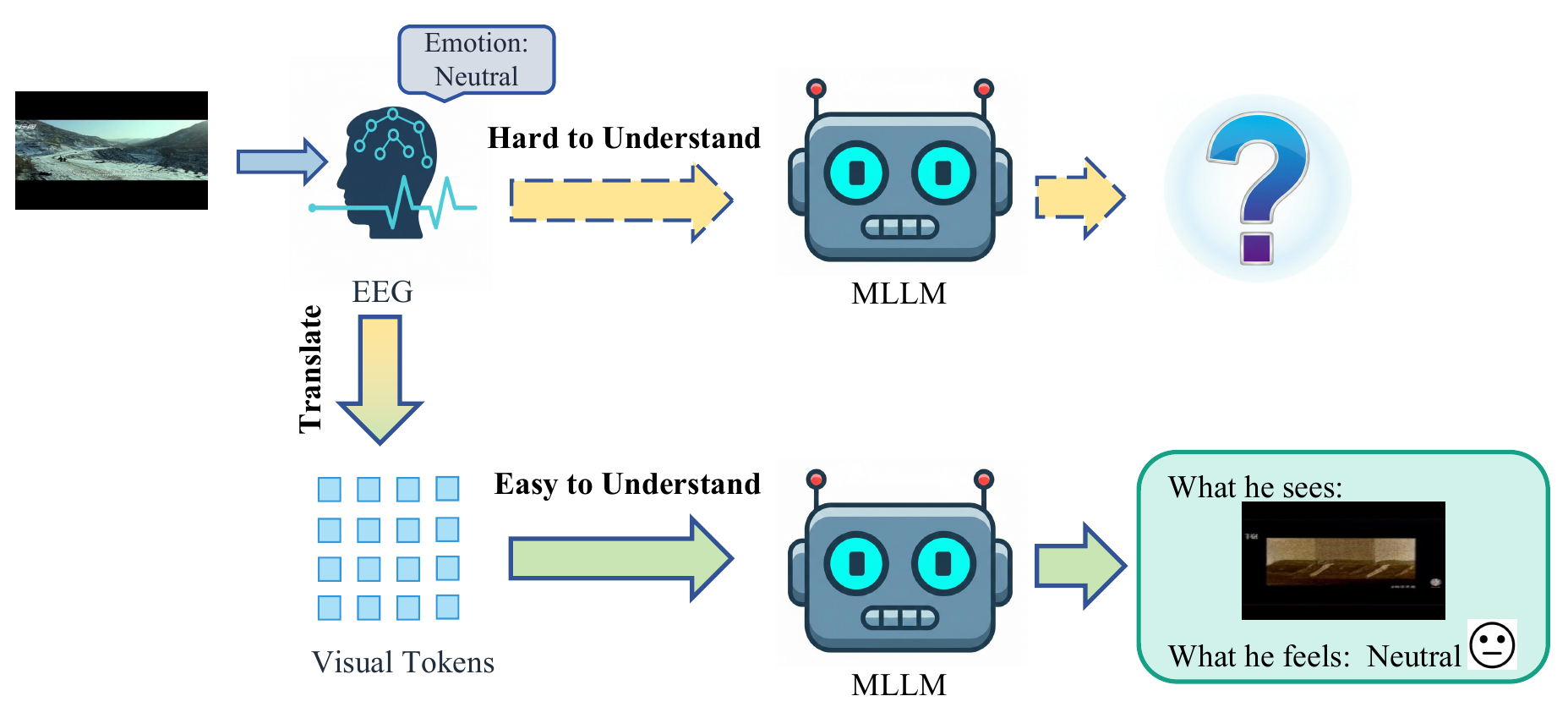}
        \caption{\textbf{Core idea of GVG.} EEG is translated into a visual-like language so that MLLMs can better interpret neural dynamics with their native visual priors.}
        \label{fig:keyidea_core}
    \end{subfigure}
    \hfill
    \begin{subfigure}{0.49\textwidth}
        \centering
        \includegraphics[width=\linewidth]{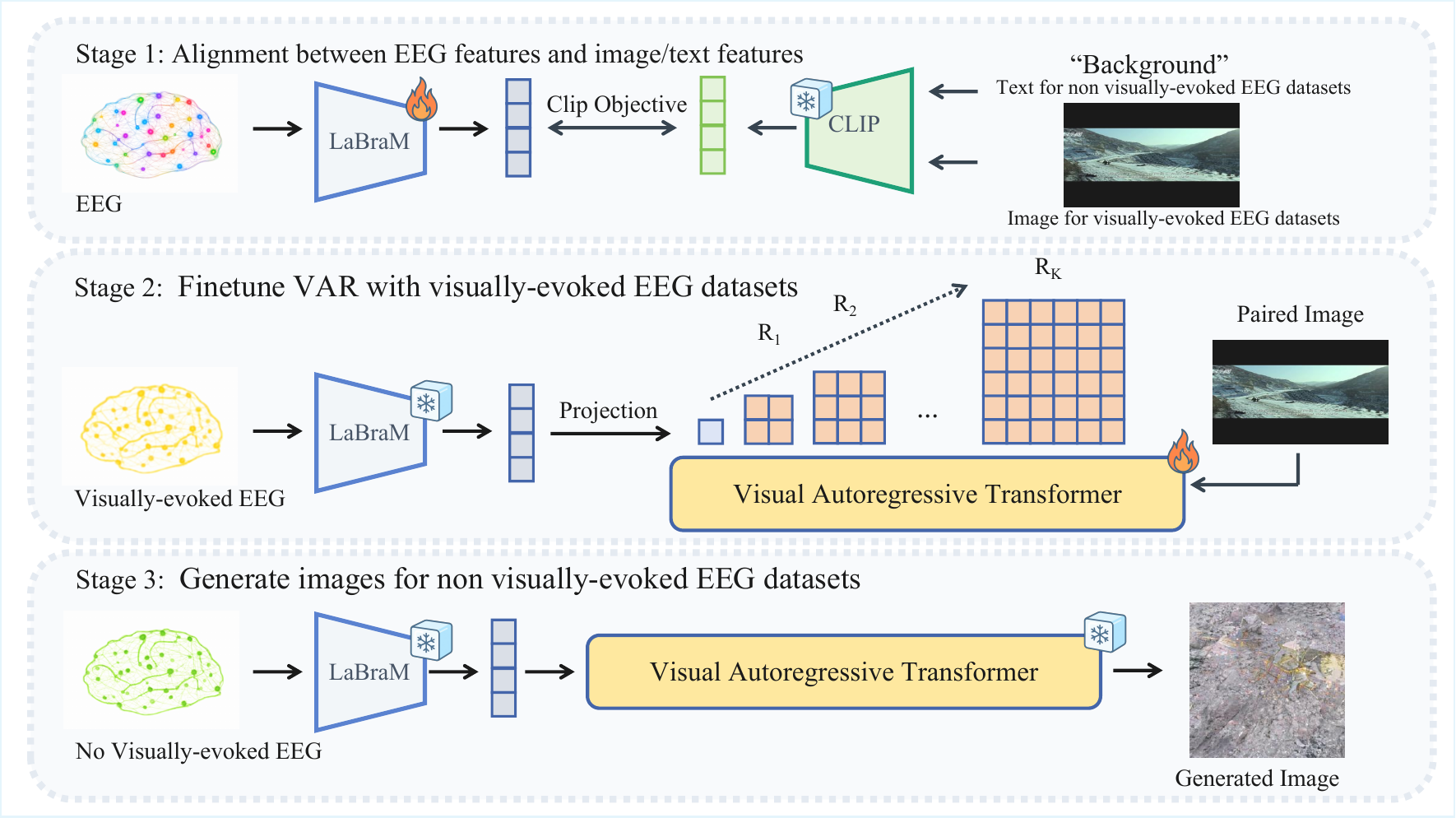}
        \caption{Training and Generation of \textbf{AVDE}~\cite{dai2026autoregressive}. For EEG datasets without paired visual stimuli, AVDE generates instance-specific proxy images to provide visual grounding.}
        \label{fig:keyidea_avde}
    \end{subfigure}
    \caption{\textbf{Overview of our core idea and proxy-image strategy.} Left: GVG converts EEG into a visual-like language, allowing MLLMs to leverage rich visual priors for understanding neural signals. Right: for non-visual clinical datasets, AVDE synthesizes proxy images that bridge the missing visual modality and enable the same visual grounding pipeline.}
    \Description{A two-panel overview figure. The left panel illustrates the core idea of translating EEG into a visual-like language for MLLMs. The right panel shows AVDE generating proxy images for non-visual datasets so that the same visual grounding framework can be applied.}
    \label{fig:keyidea}
\end{figure*}

Nevertheless, the current paradigm of compressing neural signals solely into highly abstracted textual labels is inherently lossy. While text provides explicit semantic anchors and high-level categorical boundaries, it inevitably discards the fine-grained, perceptual details and spatial topologies encoded in dense brain activity. Because human visual perception is closely coupled with complex cognitive and emotional responses, we hypothesize that augmenting text alignment with the dense visual representation space of MLLMs may provide a useful complementary signal. While textual representations provide explicit semantic anchors, the continuous visual space may offer a dense perceptual canvas. This complementarity can help preserve the fine-grained spatial topologies and low-level neural dynamics that are often weakened during pure textual compression. By combining abstract textual concepts with visual grounding, we aim to leverage the multimodal (visual-linguistic) priors encoded in MLLMs for both discriminative understanding and generative interpretability.\par
Despite the appeal of this visually-augmented paradigm, operationalizing this strategy faces a severe bottleneck: its strict reliance on paired visual stimuli. First, compared to the billions of text-image pairs used to pre-train MLLMs, visually-evoked EEG datasets are inherently scarce and small-scale. More critically, the vast majority of real-world clinical applications---such as sleep staging and epilepsy detection---are completely blind (recorded purely physiologically without any external visual input). This creates a fundamental dilemma: researchers are often forced to rely on textual alignment for broad clinical applicability, while visual alignment remains largely confined to the small subset of stimulus-driven datasets.\par
Furthermore, even if visually-evoked EEG can be aligned well, we still face a substantial scale mismatch. Unlike the massive corpora available for text or images, EEG datasets remain too limited for MLLMs to acquire fluent EEG-specific representations from scratch. To mitigate this data bottleneck, our core insight is to translate neural dynamics into a modality that MLLMs already model well: vision. By mapping raw brain signals into recognizable visual formats, we reduce the burden of learning a new neural language from limited data and instead reuse established visual priors to interpret EEG.\par
Building on this insight, we propose \textbf{Generative Visual Grounding} (GVG), a framework that visualizes the invisible. To bridge the scenario gap for purely clinical applications, we employ an EEG-to-Image generative model (e.g., AVDE~\cite{dai2026autoregressive}) as a visual translator to hallucinate instance-specific proxy images for non-visual EEG. These synthesized visual anchors allow purely physiological, non-visual datasets to be incorporated into the same visual grounding pipeline. Subsequently, to bridge the modality gap without massive parameter tuning, we map these visually grounded EEG signals into discrete image tokens. Specifically, we use the native tokenizers of off-the-shelf MLLMs (X-Omni~\cite{geng2025x} and Janus~\cite{wu2024janus}) as the interface. We train a lightweight Transformer adapter to translate aligned EEG features into the discrete codebook space of the MLLM's visual encoder, allowing the MLLM to process brain activity through its existing visual token space. These visual tokens can in turn be inspected through the corresponding visual decoder without additional tuning. To systematically validate and scale this paradigm, our exploration proceeds in two progressive stages: (1) Validation via Image-Only Grounding: We first map these visually grounded EEG signals into both discrete MLLM representation spaces using strictly image-only alignment. This step isolates the contribution of the hallucinated proxies. Under this setting, the lightweight GVG-X-Omni matches the 1.7B text-aligned baseline (NeuroLM-XL) while tuning only 170M parameters on top of a frozen 7B backbone (10 $\times$ fewer trainable parameters). (2) Maximizing Synergy via Trimodal Extension: Motivated by this image-only result, we further extend the high-capacity instantiation (GVG-Janus) with a trimodal (Image+Text) alignment. This configuration suggests a multimodal complementarity: explicit textual labels provide categorical semantic anchors, while the hallucinated visual proxies supply structural information that text alone does not preserve. In our experiments, trimodal GVG-Janus delivers the strongest overall average performance among the compared multitask models, suggesting that generative visual proxies can complement textual semantics for decoding complex neural dynamics.\par
Our main contributions are summarized as follows:
% \begin{itemize}
%     \item \textbf{Generative Visual Grounding for Blind EEG:} We propose GVG, which uses AVDE-generated proxy images to extend visual grounding to non-visual clinical EEG.
%     \item \textbf{Modality Translation via Discrete Visual Tokens:} We map EEG into the discrete visual token spaces of off-the-shelf MLLMs, enabling frozen backbones to process neural signals in a native visual interface.
%     \item \textbf{Two Practical Operating Points:} We identify a parameter-efficient setting with GVG-X-Omni and a stronger multitask setting with GVG-Janus.
%     \item \textbf{Unified Understanding and Reconstruction:} We combine EEG understanding and visual reconstruction in one framework, using reconstruction as a mechanistic probe rather than a goal of pixel-perfect synthesis.
% \end{itemize}

\begin{itemize}
    \item \textbf{Generative Visual Grounding for Blind EEG:} We propose an approach that visualizes invisible physiological states. By employing a generative model as a visual translator to hallucinate proxy images for non-visual data, we reduce the framework's reliance on stimulus-paired EEG and enable purely clinical datasets to participate in a unified visual-aligned pipeline.
    \item \textbf{Modality Translation via Discrete Visual Tokens:} By mapping raw EEG signals directly into the discrete visual codebook of off-the-shelf MLLMs, we represent noisy brainwaves in a native visual token space. This token-space translation enables frozen MLLMs to decode neural dynamics using their pre-trained visual priors.
    \item \textbf{Extreme Efficiency and Competitive Performance:} For training efficiency, GVG-X-Omni achieves performance competitive with the 1.7B text-aligned baseline (NeuroLM-XL) while tuning only 170M parameters on top of a frozen 7B backbone and using ~10\% of the pre-training data volume. For stronger cross-domain synergy, our full-capacity instantiation (GVG-Janus) incorporates hallucinated clinical images to deliver strong multitask performance and improves over NeuroLM-XL on several benchmarks at a comparable trainable-parameter scale.
    \item \textbf{Unified Understanding and visual reconstruction:} We present a unified framework that combines multi-task EEG understanding with visualization of hidden representations within a single architecture. Rather than optimizing for pixel-perfect synthesis, we use the reconstruction capability as a mechanistic probe to examine the semantic fidelity of the decoded representations.
\end{itemize}
\begin{figure*}[th]
    \centering
    \includegraphics[width=\textwidth]{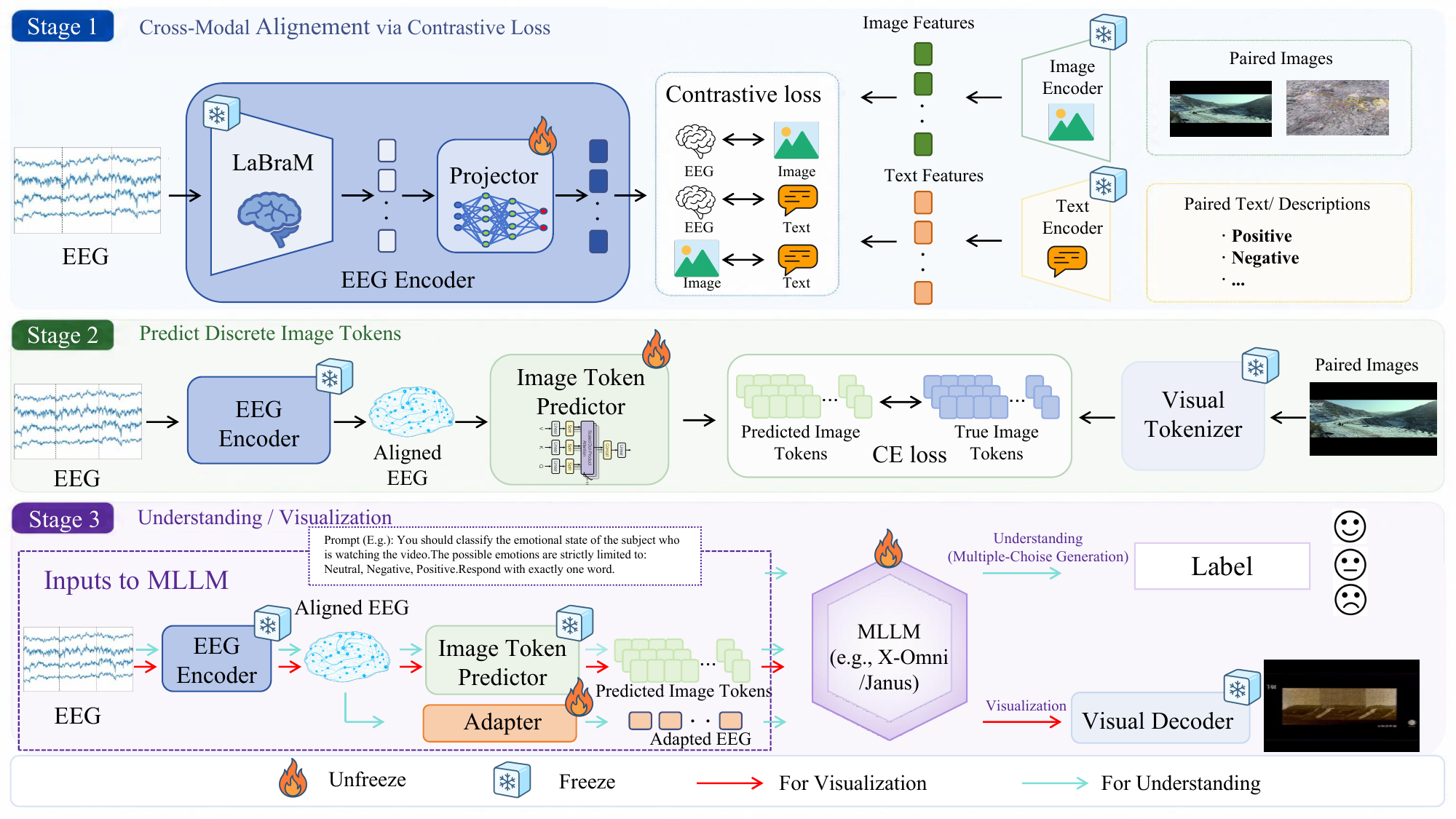}
    \caption{\textbf{Overview of the Generative Visual Grounding (GVG) Training Framework.} 
    The proposed GVG pipeline consists of three stages: cross-modal alignment, image-token prediction, and MLLM-based understanding and visual reconstruction. For non-visual clinical EEG, AVDE generates proxy images to bridge the missing visual modality. We instantiate the framework with GVG-X-Omni and GVG-Janus to study the trade-off between efficiency and multitask performance.}
    \Description{A three-stage overview of the GVG framework. Stage 1 aligns EEG with image and optionally text representations. Stage 2 predicts discrete visual tokens from aligned EEG features. Stage 3 routes those EEG-derived visual tokens into the MLLM backbone for understanding and visual reconstruction. The figure also shows AVDE generating proxy images for non-visual clinical EEG and highlights the two instantiations, GVG-X-Omni and GVG-Janus.}
    \label{fig:gvg}
\end{figure*}

\section{Related Work}
\subsection{Large Brain Foundation Models}
To transcend the limitations of subject-specific and task-narrow decoding, the field is rapidly converging towards Brain Foundation Models (BFMs). Early endeavors primarily utilized contrastive learning to align signal representations~\cite{kostas2021bendr, zhang2022self} or exploited global statistics~\cite{yang2023self, cai2023mbrain}. Moving beyond contrastive objectives, generative pre-training has become the dominant paradigm. In the realm of masked modeling, LaBraM~\cite{jiang2024large} introduced vector-quantized tokenization to enable patch-level reconstruction, a strategy further explored by BrainMAE~\cite{yang2024brainmae}, CBraMod~\cite{wang2024cbramod}, BrainWave~\cite{yuan2024brainwave}, NeurIPT~\cite{fang2025neuript}, ECHO~\cite{liu2025echo} and Uni-NTFM~\cite{chen2025uni}. Parallelly, autoregressive models have demonstrated remarkable scalability; BrainLM~\cite{caro2023brainlm} and Neuro-GPT~\cite{cui2024neuro} validated the efficacy of causal sequence modeling, while recent billion-scale models like EEGPT~\cite{wang2024eegpt, yue2024eegpt}, NeuroLM~\cite{jiang2024neurolm}, UniMind~\cite{lu2025unimind} and $E^2$-LLM~\cite{ma20262} have pushed the boundaries of multi-task generalization, following the methods of LLaVA~\cite{liu2023visual}. Furthermore, cross-modal frameworks such as Brant-X~\cite{zhang2023brant} and CEREBRO~\cite{dimofte2025cerebro} have begun to integrate EEG with diverse physiological signals. Despite these advances, most existing BFMs still learn primarily from the EEG signal itself or from relatively sparse textual supervision. How to leverage the dense visual priors of MLLMs for broader EEG understanding, especially when paired visual stimuli are unavailable, remains underexplored. This motivates our investigation of generative visual grounding as a complementary path toward universal EEG understanding.

\subsection{Cross-Modal Alignment between EEG and Vision} 
To interpret neural semantic content, research has increasingly focused on bridging the EEG-vision modality gap. Early endeavors, such as Brain2Image~\cite{kavasidis2017brain2image} and Palazzo et al.~\cite{spampinato2017deep}, utilized adversarial training to establish a mapping from neural activity to visual features. With the rise of diffusion models, works like DreamDiffusion~\cite{bai2023dreamdiffusion}, NEUROIMAGEN~\cite{lan2023seeing}, Brain-Diffuser~\cite{chen2023brain}, Mind-Video~\cite{chen2023cinematic}, EEG2Video~\cite{liu2024eeg2video}, MINDEV~\cite{huang2025mindev} and EEGMirror~\cite{liu2025eegmirror} have advanced this alignment by projecting EEG embeddings into the continuous conditional spaces of pre-trained generative models. However, these approaches primarily frame alignment as a unidirectional translation task optimized for perceptual reconstruction. Mapping EEG solely to continuous generative latents often entangles semantic concepts with low-level textures, limiting utility for discriminative tasks. In contrast, we propose to align EEG signals with \textit{discrete visual tokens} and treat the visual codebook as a universal semantic interface to translate noisy signals into structured codes for generalized understanding.

\section{Method}
In this section, we present Generative Visual Grounding (GVG), a unified framework designed to bridge the severe scenario and modality gaps between raw brainwaves and MLLMs for understanding and generation tasks.
\subsection{Preliminaries: MLLMs' Backbones}
\label{sec:mllms_preliminaries}

Our GVG framework is backbone-agnostic and can be instantiated with different MLLMs. We validate it on two architectures: X-Omni~\cite{geng2025x} and Janus-1.3B~\cite{wu2024janus}.\par
X-Omni is a unified multimodal model designed for high-fidelity image generation and understanding. X-Omni comprises three strategic components designed to bridge discrete language tokens and continuous visual signals: a semantic image tokenizer built on SigLIP2-g~\cite{tschannen2025siglip} that compresses continuous images into a sequence of discrete tokens of the codebook ($|\mathcal{V}|=16384$), a unified auto-regressive backbone based on the pre-trained Qwen2.5-7B LLM~\cite{qwen2} augmented with vision-specific blocks, and a diffusion decoder based on FLUX.1-dev~\cite{labs2025flux1kontextflowmatching} to reconstruct high-fidelity images from discrete tokens.\par
Janus-1.3B is a lightweight unified vision-language model featuring decoupled visual encoding for understanding and generation. It employs SigLIP-Large-Patch16-384\cite{zhai2023sigmoid} for understanding and a VQ-VAE tokenizer ($|\mathcal{V}|=16384$) for generation, with DeepSeek-LLM-1.3B as the language backbone.\par
Specifically, we leverage their tokenizer to ground EEG features and supply discrete ground-truth tokens. The unified auto-regressive backbone serves as the core processor for semantic understanding, while the diffusion decoder is dedicated to the visualization of the masqueraded tokens.

\subsection{Generative Visual Grounding via AVDE}
A critical bottleneck for visual-aligned EEG foundation models is the strict reliance on paired visual stimuli: purely clinical datasets recorded without external visual input cannot seamlessly participate in cross-modal visual pre-training. To overcome this fundamental limitation, we employ an EEG-to-Image generative model (AVDE~\cite{dai2026autoregressive}) as a cross-modal "visual translator" to hallucinate instance-specific proxy images for non-visual EEG data.\par
The original AVDE contains two training stages: 1) alignment between EEG features and image features; 2) finetuning VAR~\cite{tian2024visual} to generate images based on aligned EEG features. To adapt this generative capability for clinical datasets, we introduce a novel dual-modality alignment strategy during the training phase. While visually-evoked EEG is aligned with image features via a hybrid InfoNCE and MSE objective, we align the non-visual clinical EEG with CLIP text features derived from task-specific label descriptions. With the aligned EEG features, we finetune the VAR to autoregressively generate discrete VQ-VAE tokens. Because our modified Stage 1 maps textual clinical semantics into the shared CLIP space, the original visual decoder can then process these text-aligned clinical embeddings to generate meaningful visual tokens.
\subsection{Universal Multimodal Integration and Synergistic Decoding}
As illustrated in \cref{fig:gvg}, our method proceeds in three stages. First, following the paradigm of vision-language pre-training, we employ contrastive learning to align EEG representations with the visual feature space as well as text feature space. Subsequently, we introduce a Transformer-based adapter to translate these aligned features into discrete image tokens. This pivotal step converts continuous neural signals into a vision-like format, which serves as a compatible input for the MLLM backbone to execute downstream tasks and visualization.\\
\textbf{Stage 1: Cross-Modal Alignment via Contrastive Loss.}
To bridge the substantial modality gap between neural signals and visual stimuli, we employ a dual-encoder architecture trained with a pairwise contrastive objective like previous work aligning image features with text features~\cite{radford2021learning,tschannen2025siglip,zhai2023sigmoid}. As for the EEG Encoder, we utilize the pretrained \textbf{LaBraM}~\cite{jiang2024large} encoder, a large model pretrained based on EEG, which segments raw EEG into patches processed by a temporal encoder and enriched with spatiotemporal embeddings to capture global dependencies effectively. And for the visual/text encoder, we keep it the same as the corresponding MLLMs (X-Omni~\cite{geng2025x} and Janus~\cite{wu2024janus}). Given a batch of triplets $(\mathbf{x}_{eeg}^i, \mathbf{x}_{img}^i, \mathbf{t}^i)\in\mathbb{R}^{C_{eeg}\times L}\times\mathbb{R}^{C_{img}\times H\times W}\times\mathcal{T}$--- comprising continuous EEG signals ($C_{eeg}$ channels $L$), images ($C_{img}$ channels with spatial resolution $H\times W$) and task-specific textual labels from corpus $\mathcal{T}$---we extract and project all heterogeneous modalities into a shared D-dimensional latent space:
\begin{align}
    \mathbf{h}_{eeg}^i &= \phi_{eeg}(\operatorname{Enc_{eeg}}(\mathbf{x}_{eeg}^i))\in\mathbb{R}^D, \\
    \mathbf{h}_{img}^i &= \phi_{img}(\operatorname{Enc_{img}}(\mathbf{x}_{img}^i))\in\mathbb{R}^D, \\
    \mathbf{h}_{text}^i &= \phi_{text}(\operatorname{Enc_{text}}(\mathbf{t}^i))\in\mathbb{R}^D,
\end{align}
where $\phi_{eeg}$, $\phi_{img}$ and $\phi_{text}$ are learnable projectors. To achieve comprehensive cross-modal synergy, we formulate a trimodal objective that decomposes into three weighted pairwise alignment terms: 
\begin{equation}
    \mathcal{L}_{\text{tri}} = \lambda_{ei}\mathcal{L}_{ei} + \lambda_{et}\mathcal{L}_{et} + \lambda_{it}\mathcal{L}_{it},
\label{eq:trimodal}
\end{equation}
where $\lambda_{ei}$, $\lambda_{et}$ and $\lambda_{it}$ are hyper-parameters scaling the distinct cross-modal interactions. To ensure robust mapping, each pairwise loss $\mathcal{L}_{ab}$ integrates a symmetric InfoNCE contrastive term with a mean squared error (MSE) regression term:
\begin{equation}
    \mathcal{L}_{ab} = \alpha \mathcal{L}_{\text{InfoNCE}}(\mathbf{h}_a, \mathbf{h}_b) + (1-\alpha) ||\mathbf{h}_a - \text{sg}(\mathbf{h}_b)||_2^2,
\end{equation}
where $\alpha\in(0,1)$ acts as a balancing coefficient, and $\text{sg}(\cdot)$ denotes the stop-gradient operation. To rigorously isolate the efficacy of visual proxies from textual hints, our Image-only instantiations (for both GVG-X-Omni and GVG-Janus) optimize strictly the EEG-Image pairwise loss ($\mathcal{L}=\mathcal{L}_{ei}$), whereas the full trimodal objective (Eq.~\ref{eq:trimodal}) is reserved for the full GVG-Janus configuration. \par
This synergistic trimodal formulation operationalizes our core hypothesis by providing two indispensable advantages: (1) Textual Semantic Anchoring: Aligning EEG with explicit categorical text provides strict high-level semantic boundaries, which is critical for stabilizing representation learning on purely clinical datasets where hallucinated visual proxies may contain noise; and (2) Visual Perceptual Supplement: Aligning EEG with the visual space injects the fine-grained structural topologies and dense perceptual details that text inherently discards.\\
\textbf{Stage 2: Predict Discrete Image Tokens.}
While Stage 1 aligns the continuous feature spaces, X-Omni and Janus fundamentally operate on discrete tokens. To bridge this granularity gap, we translate the aligned EEG representations into the specific discrete visual codebook defined by their pre-trained image tokenizer. During training, this tokenizer processes the paired images to generate ground-truth discrete tokens, serving as the supervisory signal for our EEG decoder. 

Considering the discrepancy in visual feature spaces across different models, we adopt model-specific strategies for image token prediction. Specifically, for X-Omni, we employ a similarity-based prediction mechanism as described below, while for Janus, we utilize a learnable classification head for token prediction.

With aligned EEG representations, we employ a transformer-based image token predictor. For X-Omni, instead of enforcing rigid classification via a learnable linear head, we guide the predicted features toward the ground-truth semantics in the latent space. Concretely, for the $m$-th position, we compute the similarity between the predicted hidden state $\hat{\mathbf{h}}_m^i$ and the fixed visual codebook to derive a probability distribution:
\begin{equation}
    P(q_m = k | \mathbf{h}_{\text{eeg}}^i) = \frac{\exp(\hat{\mathbf{h}}_m^i \cdot \mathbf{v}_k / \tau)}{\sum_{j=1}^K \exp(\hat{\mathbf{h}}_m^i \cdot \mathbf{v}_j / \tau)},
\end{equation}
where $\tau$ is a temperature hyperparameter, $q_m$ is the $m$-th predicted image token, and $\mathbf{v}_k$ denotes the $k$-th embedding in the codebook $\mathcal{V}=\{\mathbf{v}_k\}_{k=1}^{K}\in\mathbb{R}^{K\times D}$.

Distinct from standard sequence generation approaches, we eschew the autoregressive paradigm in favor of a non-autoregressive (parallel) prediction strategy. We simultaneously predict the entire sequence of image tokens to avoid error accumulation and ensure robust generation.\\
\begin{table*}[t]
    \centering
    \caption{Performance comparison on five EEG benchmarks with broad public baseline coverage. We report Balanced Accuracy (B-Acc) and Weighted F1 (F1-W). Single-task models use dataset-specific heads; multi-task models share a unified architecture. \textbf{`Alignment'} denotes the cross-modal target space used during training. The best multi-task results are \textbf{bold}; the best overall are \underline{underlined}. SEED-VII is discussed separately because comparable public baselines with clearly documented evaluation splits remain sparse.}
    \label{tab:main_results}
    \resizebox{\linewidth}{!}{
    \begin{tabular}{l c c c cc cc cc cc cc}
        \toprule
        \multirow{2}{*}{Model} & \multirow{2}{*}{Multi-Task} & \multirow{2}{*}{Trainable Params} & \multirow{2}{*}{Alignment} & \multicolumn{2}{c}{SEED} & \multicolumn{2}{c}{SEED-IV} & \multicolumn{2}{c}{TUEV} & \multicolumn{2}{c}{TUAB} & \multicolumn{2}{c}{HMC} \\
        \cmidrule(lr){5-6} \cmidrule(lr){7-8} \cmidrule(lr){9-10} \cmidrule(lr){11-12} \cmidrule(lr){13-14}
         & & & & B-Acc & F1-W & B-Acc & F1-W & B-Acc & F1-W & B-Acc & F1-W & B-Acc & F1-W \\
        \midrule
        SPaRCNet       & $\times$ & 0.79M & - & 55.96 & 55.85 & 29.88 & 32.05 & 41.61 & 70.24 & 78.69 & 75.13 & 47.56 & 41.08 \\
        ContraWR       & $\times$ & 1.6M  &- & 61.06 & 61.37 & 38.38 & 40.21 & 43.84 & 68.93 & 80.17 & 80.65 & 42.42 & 29.87 \\
        CNN-Trans      & $\times$ & 3.2M  & - & 61.61 & 61.50 & 35.21 & 36.57 & 40.87 & 68.54 & 79.53 & 78.76 & 65.73 & 68.96 \\
        FFCL           & $\times$ & 2.4M  & - & 58.08 & 57.43 & 37.81 & 39.76 & 39.79 & 67.83 & 78.19 & 77.83 & 44.27 & 29.02 \\
        ST-Trans       & $\times$ & 3.5M  & - & 54.79 & 55.05 & 36.93 & 36.95 & 39.84 & 68.23 & 79.66 & 80.90 & 25.59 & 14.28 \\
        BIOT           & $\times$ & 3.2M  & - & 70.97 & 71.34 & 36.19 & 42.76 & 52.81 & 74.92 & 79.59 & 78.82 & 68.62 & 70.91 \\
        LaBraM         & $\times$ & 5.8M  & - & \underline{73.18} & \underline{73.54} & 47.63 & 49.14 & \underline{64.09} & \underline{83.12} & \underline{81.40} & \underline{81.47} & \underline{72.86} & \underline{75.54} \\
        \midrule
        NeuroLM-B      & \checkmark & 254M & Text & 55.54 & 55.99 & ---   & ---   & 45.60 & 71.53 & 78.26 & ---   & 67.37 & 71.26 \\
        NeuroLM-L      & \checkmark & 500M & Text & 60.06 & 60.48 & ---   & ---   & 41.32 & 73.87 & 78.76 & ---   & 66.58 & 68.96 \\
        NeuroLM-XL     & \checkmark & 1.7B & Text & 60.34 & 60.63 & 32.30 & 34.65 & 46.79 & 73.59 & 79.69 & 78.93 & 57.61 & 58.83 \\
        \midrule
        GVG-X-Omni (LoRA)    & \checkmark & 170M & Image & 60.73 & 61.11 & 28.05 & 22.41   & 44.15 & 71.35 & 74.45 & 74.78 & 61.20 & 63.64 \\
        GVG-Janus (Full)   & \checkmark & $\sim$1.7B & Image & 57.29 & 56.15 & 33.04 & 31.13 & 48.38 & 75.72 & 80.27 & 80.27 & 66.18 & 70.84 \\
        \midrule
        GVG-Janus (LoRA)& \checkmark & 49M  & \textbf{Image+Text} & 65.35 & 65.66 & 47.82 & 49.23 & 52.44 & 77.20 & \textbf{80.75} & \textbf{81.04} & 62.21 & 60.64 \\
        \textbf{GVG-Janus (Full)}& \checkmark & $\sim$1.7B & \textbf{Image+Text} & \textbf{68.92} & \textbf{68.49} & \underline{\textbf{62.50}} & \underline{\textbf{63.69}} & \textbf{59.81} & \textbf{82.05} & 80.20 & 80.44 & \textbf{71.61} & \textbf{73.84} \\
        \bottomrule
    \end{tabular}
    }
\end{table*}
\textbf{Stage 3: Understanding and Visualization.} In the final stage, we leverage the dual capabilities of the pre-trained MLLMs for both semantic understanding and generative visualization. Because our EEG-derived representations from Stage 2 are structurally and semantically aligned with the native visual space, they can be seamlessly routed into two distinct pathways without training any task-specific heads.\par
Understanding: Multiple-Choice Generation.We formulate classification as a constrained multiple-choice generation task. Given task-specific textual instructions and candidate labels formatted as options (A, B, C, ...), the MLLM autoregressively generates a single letter token constrained to the valid option set. We construct the input by concatenating the text prompt with both continuous EEG features and discrete predicted image tokens. Training uses teacher forcing with cross-entropy loss on the target option letter.\par
Visualization: Direct Reconstruction from Tokens.
To empirically assess whether the EEG representations are compatible with the native visual token space, we utilize visual reconstruction as a mechanistic probe. Specifically, we bypass the LLM reasoning module and directly route the predicted discrete image tokens to their corresponding visual generation backbones, interfacing with the Flux.1-dev diffusion model for GVG-X-Omni and the native tokenizer's visual decoder for GVG-Janus, without any task-specific fine-tuning. The model relies on the predicted discrete image tokens as the structural conditioning to synthesize the visual scenes. This strategy ensures that the reconstruction is generated solely from the discretized visual semantics decoded from brain signals, supporting the fidelity of the cross-modal alignment without relying on auxiliary EEG embeddings or intermediate continuous latents.
\section{Experiments}
\subsection{Details}
\label{details}
\paragraph{Datasets.}
We evaluate on six diverse EEG benchmarks spanning both visually-evoked and non-visual clinical paradigms. The visually-evoked datasets---\textbf{SEED} (3-class)~\cite{duan2013differential,zheng2015investigating}, \textbf{SEED-IV} (4-class)~\cite{8283814}, and \textbf{SEED-VII} (7-class)~\cite{10731546}---record EEG while subjects watch emotionally-evoked video clips, providing natural EEG-image pairs via central-frame alignment (4-second windows at 200 Hz, paired with the midpoint video frame). The non-visual clinical datasets have no paired visual stimuli and rely entirely on AVDE-generated proxy images for visual alignment: \textbf{TUEV}~\cite{harati2015improved} targets clinical event detection (6 types including seizure, slowing); \textbf{TUAB}~\cite{harati2015improved} is a large-scale binary abnormality detection corpus; \textbf{HMC}~\cite{alvarez2021inter} targets 5-class sleep staging (Wake/N1/N2/N3/REM). All datasets follow the same data split as in prior work to ensure fair comparisons. 
\paragraph{Training details.}
In Stage 1 (trimodal alignment), we assign reduced sampling weights to non-visual clinical datasets (TUEV: 0.3, TUAB: 0.3, HMC: 0.3 vs.\ SEED-Series: 1.0) to prevent the noisy AVDE proxy images from dominating the contrastive objective and disrupting the visual alignment learned from high-quality stimulus-paired data. In Stage 2 (image token prediction), we restrict training to visually-evoked EEG datasets, as they provide high-quality visual supervision with precise spatial details, which is critical for learning fine-grained image token representations. In Stage 3 (understanding), we apply balanced sampling across all datasets to ensure that each task contributes equally during multi-task fine-tuning, preventing large-scale datasets from overwhelming smaller ones.
\paragraph{Model configurations.}
We instantiate two complementary configurations to probe opposite ends of the efficiency-performance trade-off: \textbf{GVG-Janus} (Janus-1.3B backbone) is our full framework with trimodal alignment and AVDE proxy images. By operating at a comparable parameter scale to NeuroLM (1.7B), this configuration is designed to test whether generative visual grounding can remain competitive with strong text-aligned baselines under similar parameter budgets in a unified multitask setting. \textbf{GVG-X-Omni} (X-Omni/Qwen2.5-7B backbone, $\sim$170M trainable parameters, 7B frozen): our lightweight instantiation using pairwise image alignment and the massive frozen X-Omni backbone. By freezing the 7B-parameter LLM and training only $\sim$170M adapter parameters, this configuration demonstrates the \textit{trainable-parameter efficiency} of our approach---matching billion-parameter text-aligned baselines (NeuroLM) with a fraction of the trainable parameters and pre-training data.

\subsection{Performance on Understanding Tasks}
To rigorously evaluate our framework, we benchmark our models against established specialist models, including SpaRCNet~\cite{jing2023development}, ContraWR~\cite{yang2023self}, CNN-Transformer~\cite{peh2022transformer}, FFCL~\cite{li2022motor}, ST-Transformer~\cite{song2021transformer}, BIOT~\cite{yang2023biot}, LaBraM~\cite{jiang2024large}, and unified multi-task EEG models, NeuroLM~\cite{jiang2024neurolm}. The results are detailed in \cref{tab:main_results}. It is important to note that the specialist baselines use separate dataset-specific classification heads and are optimized independently for each task, whereas our model performs prompt-based multi-task understanding with a single shared MLLM backbone across all datasets. This unified setting is inherently more constrained, yet GVG-Janus still outperforms most specialist models, remains close to LaBraM overall, and even surpasses LaBraM on SEED-IV. We focus Table~\ref{tab:main_results} on the five benchmarks with broad public coverage. For SEED-VII, comparable public baselines remain sparse and some recent reports do not document sufficiently detailed evaluation splits, so we discuss it separately in the alignment ablation and analysis sections rather than mixing sparse and dense baseline coverage in a single table.\\
\textbf{Efficacy of Image-Only Visual Grounding.} To isolate the effect of visual grounding without the influence of textual supervision, we first evaluate our framework under an image-only grounding configuration. Both our instantiations (GVG-X-Omni and GVG-Janus) map raw EEG signals into discrete visual image tokens, which are subsequently embedded and fed into the MLLM backbones via Image-only alignment. 

It is important to contextualize the asymmetry in this comparison: the state-of-the-art NeuroLM relies on massive-scale pre-training across 25,000 hours of EEG data, whereas our framework utilizes merely $\sim$2,500 hours (10$\times$ less). Despite this data disadvantage, our lightweight GVG-X-Omni matches NeuroLM-XL (1.7B) on the SEED dataset. One plausible explanation for this efficiency is the bottleneck induced by discrete visual tokenization. Mapping continuous neural dynamics into a pre-trained discrete codebook may suppress part of the high-frequency variability and turn the alignment problem into a more structured token prediction task.  

While GVG-X-Omni is attractive from the perspective of trainable-parameter efficiency, it still relies on a frozen 7B-parameter backbone, which increases total memory use and inference cost. By contrast, GVG-Janus operates at a smaller overall scale (~1.7B total parameters). Evaluated under the same Image-only setting, GVG-Janus maintains competitive representational capacity across domains. Considering this trade-off between performance and deployment cost, we use GVG-Janus as the main backbone for the subsequent multimodal experiments.\\
\textbf{Maximizing Synergy via Trimodal Extension.} 
Given this trade-off, we next examine the benefits of extending GVG-Janus with a Trimodal (Image+Text) alignment. As shown in Table~\ref{tab:main_results}, this multimodal setting yields the strongest average performance among our multitask configurations and improves over the text-aligned NeuroLM baselines on several benchmarks. GVG-Janus provides two complementary operating points. In a parameter-efficient regime, GVG-Janus-LoRA leverages both textual anchors and visual details to substantially reduce the learning difficulty. With only 49M trainable parameters, it improves over NeuroLM-XL on SEED, SEED-IV, TUEV, and TUAB, although it remains below the strongest task-specific specialists on several datasets. At full capacity, trimodal GVG-Janus achieves 71.61\% on HMC and delivers the strongest average performance among the compared multitask models. We hypothesize that the degradation of larger text-only models on physiological datasets is related to the low information entropy of sparse label supervision, whereas trimodal supervision---combining explicit textual anchors with dense visual topologies---stabilizes scaling and improves robustness on more complex tasks.
% \textbf{Maximizing Synergy via Trimodal Extension.} 
% Given this trade-off, we next examine the benefits of extending GVG-Janus with a Trimodal (Image+Text) alignment. As shown in Table~\ref{tab:main_results}, this multimodal setting yields the strongest average performance among our multitask configurations and improves over the text-aligned NeuroLM baselines on several benchmarks. GVG-Janus provides two complementary operating points:
% \begin{itemize}
%     \item \textbf{Parameter-Efficient Trimodal Synergy (GVG-Janus-LoRA):} Armed with both textual anchors and visual details, the learning difficulty is reduced substantially. With only 49M trainable parameters, GVG-Janus-LoRA improves over NeuroLM-XL on SEED, SEED-IV, TUEV, and TUAB, although it remains below the strongest task-specific specialist on several datasets. This result suggests that trimodal grounding can materially strengthen a unified multitask model even in a parameter-efficient setting.
%     \item \textbf{Scaling with Balanced Multimodal Supervision (GVG-Janus Full):} At full capacity, trimodal GVG-Janus reaches 71.61\% on HMC and achieves the strongest average performance among the compared multitask models. We hypothesize that the degradation of larger text-only models on physiological datasets is related to the low information entropy of sparse label supervision; in contrast, trimodal supervision combines explicit textual anchors with dense visual topologies, which appears to stabilize scaling and improve robustness on more complex tasks.
% \end{itemize}
\begin{table}[t]
    \centering
    \caption{\textbf{Visual Reconstruction Quality.} We compare AVDE (baseline proxy generator) against our two GVG instantiations. \textbf{Bold}: best; \underline{underlined}: second best.}
    \label{tab:gen_metrics}
    \resizebox{\linewidth}{!}{
    \begin{tabular}{ll ccc}
        \toprule
        \textbf{Dataset} & \textbf{Method} & \textbf{PSNR} $\uparrow$ & \textbf{SSIM} $\uparrow$ & \textbf{LPIPS} $\downarrow$ \\
        \midrule
        \multirow{3}{*}{SEED} 
        & AVDE          & 10.17 & \textbf{0.3162} & \textbf{0.5072} \\
        & GVG-X-Omni    & \textbf{10.69} & 0.2943 & 0.7201 \\
        & GVG-Janus     & \underline{10.52} & \underline{0.2989} & \underline{0.6372} \\
        \midrule
        \multirow{3}{*}{SEED-IV} 
        & AVDE          & 12.47 & 0.4392 & \underline{0.5500} \\
        & GVG-X-Omni    & \underline{14.80} & \underline{0.5179} & 0.6176 \\
        & GVG-Janus     & \textbf{15.15} & \textbf{0.5767} & \textbf{0.4804} \\
        \midrule
        \multirow{3}{*}{SEED-VII} 
        & AVDE          & 11.78 & 0.3496 & \underline{0.6102} \\
        & GVG-X-Omni    & \textbf{14.41} & \textbf{0.4335} & 0.6700 \\
        & GVG-Janus     & \underline{13.48} & \underline{0.4149} & \textbf{0.5751} \\
        \bottomrule
    \end{tabular}
    }
\end{table}
\section{Analysis}
\subsection{Qualitative and Quantitative Visual Verification}
\begin{figure*}[t]
    \centering
    \begin{minipage}{0.48\linewidth}
        \centering
        \includegraphics[width=\linewidth]{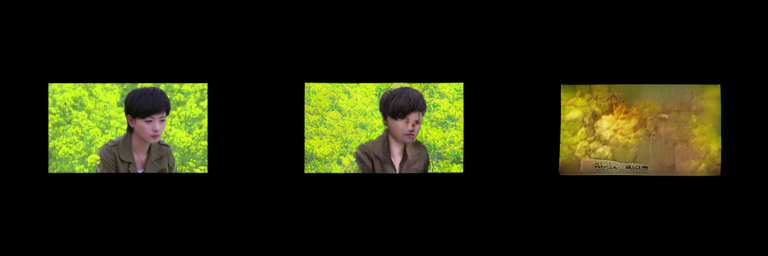} 
        \centerline{\small (a) GVG-Janus: Meadow Scenario}
    \end{minipage}
    \hfill
    \begin{minipage}{0.48\linewidth}
        \centering
        \includegraphics[width=\linewidth]{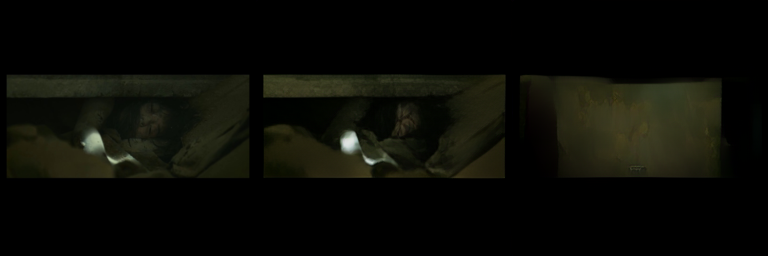}
        \centerline{\small (b) GVG-Janus: Dark Confined Scenario}
    \end{minipage}
    
    \vspace{0.2cm}
    
    \begin{minipage}{0.48\linewidth}
        \centering
        \includegraphics[width=\linewidth]{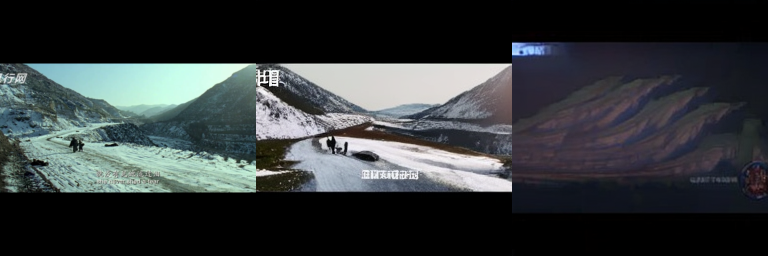}
        \centerline{\small (c) GVG-X-Omni: Snowy Mountain Scenario}
    \end{minipage}
    \hfill
    \begin{minipage}{0.48\linewidth}
        \centering
        \includegraphics[width=\linewidth]{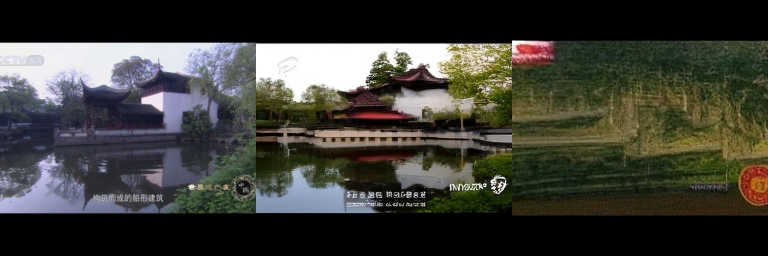}
        \centerline{\small (d) GVG-X-Omni: Traditional Pavilion Scenario}
    \end{minipage}
    
    \caption{\textbf{Qualitative Results of EEG-based Visual Reconstruction.} We visualize the decoding capabilities of our two instantiations. Each subfigure displays three columns: \textbf{Left:} Original Visual Stimulus; \textbf{Middle:} Ground-Truth (GT) Token Reconstruction; \textbf{Right:} EEG-Predicted Reconstruction. Despite the inherent noise in non-invasive EEG, both GVG-X-Omni and GVG-Janus recover the global topological skeleton, color palettes, and semantic atmosphere of the original scenes.}
    \Description{A four-panel qualitative comparison of EEG-based visual reconstruction. Each panel shows an original visual stimulus, its ground-truth token reconstruction, and the image reconstructed from EEG-predicted tokens. Panels (a) and (b) correspond to GVG-Janus examples, while panels (c) and (d) correspond to GVG-X-Omni examples. The figure highlights recovery of coarse scene structure, color palette, and overall atmosphere rather than fine-grained texture.}
    \label{fig:visual_reconstruction}
\end{figure*}
Utilizing visual decoders as mechanistic probes, we visualize the learned representations associated with neural dynamics. To provide a like-for-like evaluation of open-ended, class-unconditional EEG-to-Image generation, we benchmark against the state-of-the-art AVDE, as both methods reconstruct images solely from EEG features without auxiliary categorical text prompts.\\
\textbf{Quantitative Analysis.} Table~\ref{tab:gen_metrics} shows that as task complexity scales (SEED to SEED-VII), our GVG instantiations exhibit consistent advantages. While AVDE occasionally achieves better LPIPS scores (favoring high-frequency textures), GVG consistently dominates in PSNR and SSIM. This reveals a critical paradigm shift: AVDE tends to hallucinate photorealistic but structurally mismatched images, whereas GVG faithfully preserves pixel-level spatial layouts and global structural topologies by anchoring EEG to discrete visual tokens.\\
\textbf{Qualitative Insights.} Figure~\ref{fig:visual_reconstruction} illustrates this structural preservation. While fine-grained micro-textures are naturally attenuated due to EEG's low spatial resolution, GVG still recovers coarse semantic structure and atmospheric cues. For instance, GVG-Janus recovers dominant color palettes and global illumination in the Meadow and Dark Confined scenarios. Similarly, GVG-X-Omni captures the diagonal ridge-like topology in the Snowy Mountain and the horizontal spatial layout in the Pavilion. These reconstructions suggest that GVG can translate part of the latent perceptual content in EEG into interpretable visual patterns.\\
\textbf{Cognitive Plausibility \& Limitations.} We acknowledge that the synthesized images lack high-frequency micro-details. However, reconstructing a single static snapshot from a continuous EEG window of a subject watching dynamic, emotion-eliciting videos is an inherently ill-posed problem. One possible interpretation is that this abstraction is consistent with the coarse gist-level information emphasized during naturalistic video perception, such as global atmosphere, primary hues, and broad spatial boundaries rather than pixel-perfect detail. Under this view, the reconstructions are better interpreted as coarse visual summaries reflected in the EEG window, rather than literal physical reconstructions of internal percepts.

\subsection{Alignment Strategy}
\label{sec:alignment_ablation}

\begin{table}[t]
    \centering
    \caption{Ablation of Stage 1 alignment strategy. We compare Image-only (EEG$\leftrightarrow$Image), Text-only (EEG$\leftrightarrow$Text), and Trimodal (EEG$\leftrightarrow$Image$\leftrightarrow$Text) alignment. Balanced Accuracy is reported.}
    \label{tab:alignment_ablation}
    \resizebox{\linewidth}{!}{
    \begin{tabular}{l cccccc c}
        \toprule
        Alignment & SEED & SEED-IV & SEED-VII & HMC & TUAB & TUEV & Avg \\
        \midrule
        Text-only    & 54.85 & 25.63 & 19.03 & 42.89 & 80.11 & 44.41 & 49.96 \\
        Image-only   & 57.29 & 33.04 & 26.64 & 66.18 & 80.27 & 48.38 & 51.97 \\
        \textbf{Trimodal} & \textbf{68.92} & \textbf{62.50} & \textbf{34.87} & \textbf{71.61} & \textbf{80.20} & \textbf{59.81} & \textbf{62.99} \\
        \bottomrule
    \end{tabular}
    }
\end{table}

To validate our core hypothesis regarding modality complementarity, we conduct a comprehensive ablation study on the Stage 1 alignment strategy. As presented in Table~\ref{tab:alignment_ablation}, we compare models trained solely with EEG-Text alignment, solely with EEG-Image alignment, and our proposed synergistic Trimodal (EEG$\leftrightarrow$Image$\leftrightarrow$Text) alignment. The results expose a profound hierarchy in cross-modal representation learning.\\
\textbf{The Insufficiency of Single-Modality Alignment.} 
The \textit{Text-only} alignment yields the lowest average performance. While it performs adequately on coarse binary tasks, it suffers severe performance degradation on complex, fine-grained tasks. This empirically confirms that abstract text labels possess low information content, forcing the model to memorize noise rather than learning generalizable neural dynamics. Conversely, the \textit{Image-only} alignment provides a notable boost, particularly on HMC. By providing dense structural topologies, the visual space prevents overfitting. However, lacking explicit categorical guidance, the model struggles to draw sharp decision boundaries in the highly complex visual space.\\
\textbf{The Power of Multimodal Synergy (1+1 > 2).} 
The \textbf{Trimodal} alignment substantially exceeds the two single-modality variants, achieving an average accuracy of 62.99\%. The gains are strongest on highly complex tasks: on SEED-IV, the accuracy rises from 33.04\% (Image-only) to 62.50\%. On TUAB, however, Trimodal and Image-only are essentially tied, suggesting that this coarse binary abnormality task already benefits strongly from visual grounding alone and leaves limited headroom for additional text anchoring. \\
\textbf{Mechanistic Insights.} 
We attribute these gains to the complementary nature of the two target spaces. In the trimodal framework, the textual modality acts as a robust semantic anchor, explicitly guiding the MLLM toward clear categorical boundaries. Simultaneously, the visual modality acts as a dense perceptual canvas, supplying the fine-grained spatial topologies and macro-level structural regularizers that text inherently discards. Furthermore, for purely clinical, "blind" datasets (HMC, TUEV), the consistent improvements indicate that AVDE-generated proxy images provide useful complementary supervision when combined with clinical text descriptions. We incorporate textual supervision into AVDE training to constrain proxy images to encode meaningful semantics, which facilitates better alignment between EEG features and the visual space. Experimental results verify the effectiveness of this design, showing that semantic-aware proxy generation improves both alignment quality and downstream performance, suggesting that visual grounding and textual alignment are complementary.\\
\begin{table}[t]
    \centering
    \caption{Stage-wise ablation of the GVG pipeline. We compare removing Stage 1 (cross-modal alignment), removing Stage 2 (discrete image-token prediction), and the full pipeline. Balanced Accuracy is reported on the six benchmarks used throughout the main paper.}
    \label{tab:stage_ablation}
    \resizebox{\linewidth}{!}{
    \begin{tabular}{l cccccc c}
        \toprule
        Setting & SEED & SEED-IV & SEED-VII & HMC & TUAB & TUEV & Avg \\
        \midrule
        w/o Stage 1 & 34.60 & 24.67 & 14.57 & 30.58 & 67.23 & 20.75 & 32.07 \\
        w/o Stage 2 & 63.41 & 48.23 & 31.36 & 59.58 & 79.87 & 50.76 & 55.54 \\
        \textbf{Full Pipeline} & \textbf{68.92} & \textbf{62.50} & \textbf{34.87} & \textbf{71.61} & \textbf{80.20} & \textbf{59.81} & \textbf{62.99} \\
        \bottomrule
    \end{tabular}
    }
\end{table}\par
To further test whether the gains arise from the full three-stage pipeline rather than from the backbone alone, we conduct a stage-wise ablation in Table~\ref{tab:stage_ablation}. Removing Stage 1 causes performance to collapse on the multi-class emotion benchmarks, even close to chance level. This result indicates that the MLLM backbone cannot directly exploit raw EEG without an explicit cross-modal bridge. Removing Stage 2 leads to a smaller but still consistent degradation, reducing the average balanced accuracy from 62.99\% to 55.54\%. The drop is especially visible on SEED-IV, HMC, and TUEV, suggesting that Stage 2 further refines the aligned EEG features into a tokenizer-compatible space that benefits both downstream understanding and the reconstruction pathway. Together, these results support a simple division of labor: Stage 1 is necessary to establish usable cross-modal alignment, while Stage 2 provides an additional and stable refinement of the aligned representation.

\section{Conclusion}
In this work, we introduced Generative Visual Grounding (GVG), a framework for integrating continuous brain signals into Multimodal Large Language Models (MLLMs) through generated visual proxies. Rather than relying solely on textual mapping or requiring paired visual stimuli in clinical settings, we use an intermediate generative model to synthesize surrogate visual representations for arbitrary EEG data. This design provides a practical way to connect non-visual EEG with the visual priors of MLLMs.

Our systematic evaluations yielded two main takeaways for the BCI community. First, mapping neural signals to the visual modality can be effective and parameter-efficient in our setting. Our image-only, discrete instantiation (GVG-X-Omni) achieves competitive performance against the 1.7B text-aligned baseline while tuning ~170M parameters on top of a frozen 7B backbone (10$\times$ fewer trainable parameters). Second, multimodal complementarity appears more effective than either text-only or image-only supervision alone within our current framework. By combining explicit text labels with hallucinated visual proxies, GVG-Janus achieves the strongest overall average among the compared multitask models and improves over NeuroLM-XL on several benchmarks.

Overall, GVG provides a promising blueprint for developing more universal, scenario-agnostic brain foundation models. Our results support the view that generative visual translation can be a useful complement to textual alignment, while also highlighting the need for further controls to isolate which parts of that gain come specifically from visual proxy semantics.

\bibliographystyle{ACM-Reference-Format}


\end{document}